\RequirePackage{fix-cm}

\documentclass[smallextended]{svjour3}       
\usepackage{slashbox}
\usepackage{graphicx}
\usepackage{amstext}
\usepackage{adjustbox}

\usepackage{mathptmx}
\usepackage{amsmath}

\usepackage{subfig}

\usepackage{latexsym}
\usepackage[misc]{ifsym}
\usepackage{algorithmic,algorithm}
\usepackage{float}
\usepackage{calc}

\usepackage{url}
\usepackage{amsmath,amssymb,amsfonts}
\usepackage[flushleft]{threeparttable}
\usepackage{caption, multirow, makecell}
\usepackage[utf8]{inputenc}
\usepackage{diagbox}
\setcellgapes{3pt}

\begin{document}

\title{Using Context Information to Enhance Simple Question Answering}

\author{Lin Li \and
     Mengjing Zhang \and
    Zhaohui Chao \and Jianwen Xiang
}


\institute{ Lin Li \and Mengjing Zhang \and Zhaohui Chao \and Jianwen Xiang\at
              School of Computer Science and Technlogy, Wuhan University of Technology, Wuhan, China\\
           \and
           Lin Li \\
           \email{cathylilin@whut.edu.cn}\\
           \and
           Mengjing Zhang \\
              \email{zhangmengijng@whut.edu.cn}\\
           \and
           Zhaohui Chao \\
              \email{chaozhaohui@whut.edu.cn}\\
           \and
           Jianwen Xiang\\
              \email{xiangjw@gmail.com}\\
}

\date{Received: date / Accepted: date}

\maketitle

\begin{abstract}
With the rapid development of knowledge bases (KBs), question answering (QA) based on KBs has become a hot research issue. In this paper, we propose two frameworks (i.e., a pipeline framework, an end-to-end framework) to focus on answering single-relation factoid questions. In both of two frameworks, we study the effect of context information on the quality of QA, such as the entity's notable type, out-degree. In the pipeline framework, it includes two cascaded steps: entity detection and relation detection. In the end-to-end framework, we combine char-level encoding and self-attention mechanisms, using weight sharing and multi-task strategies to enhance the accuracy of QA. Experimental results show that context information can get better results of simple QA whether it is the pipeline framework or the end-to-end framework. In addition, we find that the end-to-end framework achieves results competitive with state-of-the-art approaches in terms of accuracy and take much shorter time than them.
\end{abstract}

\section{Introduction}
QA is a classic natural language processing (NLP) task, which aims at building systems that automatically answer questions formulated in natural language \cite{b1}.

In recent years, several large-scale general purpose KBs have been constructed, including Freebase \cite{b2,b3}, its main data source is Wikipedia, and some data comes from IMDB and other websites \cite{b4}, YAGO \cite{b5}, DBpedia \cite{b6} and Wikidata \cite{b7}, Babelnet \cite{b8}. Also, the open Chinese KBs include Zhishi.me \cite{b9}, CN-DBpedia \cite{b10}, Xlore \cite{b11}, etc. In addition, there are some commercial KBs that are not completely open, such as Google Knowledge Graph \cite{b4,b12,b13} and the Facebook Graph. We can access the data of entities and relations through a specific API. Since Google put forward the concept of knowledge graph, the related research of KBs has reached a new level of popularity.

The father of the World Wide Web Berners-Lee \cite{b14} proposed the concept of semantic web in 1998. Different from traditional Web networks, semantic and structured descriptions are introduced in the Semantic Network, which realizes semantic-based associations between data. And it can promote the development of computers in the direction of semantic-based information exchange \cite{b15}. RDF (Resource Description Framework) is a simple and flexible data model used to describe the semantic network proposed by the World Wide Web Consortium (W3C) \cite{b16}. Most of KBs store information in the form of RDF triples (subject, predicate, object) \cite{b16,b17,b18}. For example, (/m/02mjmr, /people/person.place\_of\_birth,  /m/02hrh0\_), where \textbf{\emph{/m/02mjmr}} is the Freebase id for Barack Obama, and \textbf{\emph{/m/02hrh0\_}} is the id for Honolulu. By structuring knowledge stored in this basic form, we can better organize, manage, and utilize vast amounts of knowledge. But people cannot directly understand and extract the knowledge in the knowledge graph without struct query language. Therefore by mapping the questions to the triples in the KBs, we can get the correct answer. This is a good way to use the KBs. With the development of KBs, question answering over knowledge base (KB-QA) \cite{b19} has attracted more and more attention.

KB-QA is defined as the task of retrieving the correct entity or a set of entities from a KB given a query expressed as a question in natural language \cite{b1}. For instance, in order to answer the question \textbf{\emph{``where was former U.S. President Obama born? "}} we need to retrieve the entity \textbf{\emph{/m/02mjmr}} in Freebase to represent former U.S. president Barack Obama and the relation \textbf{\emph{/people/person.place\_of\_birth}} corresponds to the relation of this question. With the entity and the relation, we can form a corresponding structured query. As a result, we can obtain the right answer \textbf{\emph{Honolulu}} (id: /m/02hrh0\_). Typically, with SPARQL (an RDF query language) people can extract information expressed in natural language from KBs. Just like cross-modal retrieval \cite{b20}.

The KB-QA technology can be divided into two technical routes: (1) symbol-based representations, such as traditional semantic parsing, and (2) distribution-based embedding. The former transforms the semantics of the question into a structured query, and then returns the answer by the structured query, such as $\lambda$ paradigm \cite{b21} and DCS-Tree \cite{b22}, The corresponding semantic parsing methods are: combined category grammar (CCG) \cite{b23} and dependent combination grammar (DCS) \cite{b24}. The method is efficient. However, in this process, experts are required to formulate dictionaries or grammar rules etc. Not only does it need a large amount of  human resources, it is also difficult to migrate \cite{b25}. In the latter route, the candidate answers are first determined in the KB based on the question. The question and the candidate answers are mapped to a low-dimensional space, therefore their distributed representation (i.e., Distributed Embedding) is obtained, which is trained by the training data, therefore the question vector and its corresponding correct answer vector are in a low-dimensional space. The matching score is as high as possible. When the training of the model is completed, it can be screened according to the score of the vector expression of the candidate answer and the question, and the highest score is found as the final answer. This route can be divided into a pipeline framework and an end-to-end framework. It has strong operability and does not require any manual features. Therefore its improvement space is relatively large.

With the emergence of deep learning, the development of NLP has greatly promoted. The effect of KB-QA can be improved by combing deep learning with the above two technical routes respectively. In this paper, we mainly discuss the impact of  the second route (i.e., Distribution Embedding) combined with deep learning.

Unlike traditional methods, deep learning can capture the semantic information of the text at multiple levels through learning, including words, phrases, sentences, paragraphs and chapters. Therefore the semantic gap in traditional NLP is improved or solved to some extent. Questions expressed in natural language can be answered directly. The core idea of deep learning system in KB-QA tasks is representation and matching. First, we should learn representation of both the question and the fact of KBs, which contains literature level and semantic level. Then we need to calculate the correlation between question and fact. The work of answering single-relation factoid questions was first proposed by Bordes et al. \cite{b26}. In this work, they employ Memory Networks and introduce a new dataset SimpleQuestions, which is built on Freebase. For each triple (entity1, relation, entity2), it shows the relationship between entity1 and entity2 \cite{b27}. The relation consists of entity's type and properties, defined by the Freebase ontology \cite{b17}. There are currently two main frameworks for solving KB-QA, pipeline frameworks( \cite{b28,b29,b30,b31}) and end-to-end frameworks( \cite{b1,b26,b32,b33,b34,b35,b36,b37,b38}) respectively. As in the latter case, the data processing procedure is more concise, so it is more popular than the former.

In this paper, we focus on answering single-relation factoid questions (i.e., simple questions), which can be answered by a single fact of the KBs. The task is also called simple QA. Thereby we explore two different frameworks to answer such questions, analyzing the impact of entity's context information to answer selection. We find that combining the entity's out-degree and notable type can improve the accuracy of answers. Moreover, our end-to-end framework that combines context information achieves a competitive result with state-of-the-art work of Lukovnikov et al. \cite{b34}, but run much faster than it.

In our pipeline framework and end-to-end framework, there are different ways to integrate entities' type information and out-degree information.

In our pipeline framework, For the combination of out-degree information, after the entity candidate set and the relation with the highest matching degree are obtained, in order to solve the case where multiple entities in the entity candidate set have the same relation, these entities are reordered by using the out-degree information to generate a final result. For the incorporation of type information of the entity, we calculate the matching score of the question-type pair to mine the entity type information in the question, and sort the entity candidate set and the relation candidate set to achieve final screening of the results. In the following we will show it in details.

In our end-to-end framework, type information is incorporated into two different methods. The first is to concatenating the entity type information with the entity tag and inputting it as a new tag of the entity into the multitasking end-to-end model (QA-T) to sort the answers. The second is to add a matching task between the entity type information and the question on the basis of QA-T, the entity type information is added to the original input layer, through the madel we can obtain the semantic matching score between the type information and the question, then we sort the answers based on matching scores to get the final answer. In addition, for the out-degree information of the entity, it is used to sort the answers with the highest matching score, so as to further sort and filter the answers. The detailed method will be described later.

The experimental results show that the accuracy of both frameworks is improved after combining the context information. What's more, in our end-to-end framework, combining type information, using char-level embedding and self-attention mechanisms, model QA-T can get better results. As is known, the work done by Lukovnikov et al \cite{b34} using end-to-end framework gets  71.2\% accuracy, and takes 48 hours. However, our work achieves 71.8\% but only takes 4 hours. The result is a little better, and the time spent is greatly shortened.

The rest paper is structured as follows: We introduce our related work in Section \ref{Related work}. In Section \ref{Approach} we present our two different frameworks in details. The experiment is reported in Section \ref{Experiments}, and then we conclude in Section \ref{Conclusion} with some discussions for future work.

\section{Related Work}\label{Related work}

With the development of deep learning technology, there are two mainstream technical routes for the KB-QA task: pipeline frameworks and end-to-end frameworks.

\subsection{Pipeline Frameworks}
In pipeline frameworks, the KB-QA task is decomposed into into two subtasks: entity detection and relation detection. Dai et al. \cite{b28} use the bidirectional cyclic neural network (i.e., Bi-GRU-CRF) combined with the conditional random field (i.e., CRF) to identify the questions, and cut the entity candidate sets according to the entity identification results. They also use a Bi-GRU network to encode relations in relation matching. Yin et al. \cite{b29} divide the task into two parts: entity linking and fact selection. they use the convolutional neural network based on character level (char-level) and word level (word-level) to perform entity matching and relation matching respectively, and they use the pooling method combining attention mechanism in the relation matching process. Cui et al. \cite{b31} design a new kind of question representation: templates, and use them to improve the semantic representation of questions and better judge the type and intent of questions. Moreover, they increase the coverage of the KB by 57 times through expanding predicates in the RDF KB.  Hao et al. \cite{b30} use Bi-LSTM-CRF to identify the entities in questions, then the question templates are used to correct the entity recognition results, and the relation detection is combined to improve fact selection. Multi-granular coding and multi-dimensional information are utilized in this framework.

\subsection{End-to-end Frameworks}
In end-to-end frameworks, Bordes et al. \cite{b1} present the question and fact triples in the KBs as numerical vectors of the same semantic space by embedding model, and the similarity of their vectors is measured to sort the candidate triples. Yih et al. \cite{b38} build a semantic matching model using convolutional neural networks, and propose a QA framework for single-relation questions based on semantic similarity. By measuring the entity text and entity labels, relation templates and relations in the question respectively, the similarity between the two completes the sorting of the candidate answers. Golub and He \cite{b33} employ a char-level, attention-based encoder-decoder framework for QA. The model is robust for unseen entities since it adopts char-level modeling. Hao et al. \cite{b32} present an end-to-end neural network model to represent the questions, which improves the representation of questions via cross-attention mechanisms. Lukovnikov et al. \cite{b34} present an end-to-end neural network. They merge word-level and char-level representation of questions. Wu et al. \cite{b35,b36,b37} employ attention mechanisms and joint learning. And they design an end-to-end network structure.

Although the above two frameworks have achieved good results on QA, they have not used context information for in-depth analysis to improve the accuracy of QA. Therefore, in this paper, we mainly discuss the impact of context information on the accuracy of QA.

\subsection{Attention Mechanism}
The attention mechanism is derived from the visual field. When the human eyes observe an image, it is preferred to quickly scan the entire image first and find the areas in the image that need to be focused, then they focus on these areas, carefully observe and analyze, and get more detailed information. The attention mechanism can be divided into Hard Attention, Soft Attention, Local Attention and Global Attention. Xu et al. \cite{b39} divide attention into Soft Attention and Hard Attention in dealing with image description generation tasks. The former refers to assigning weights to all regions of the original image. Part of the original image is assigned a corresponding weight. The latter can reduce the amount of calculations compared to the former. Relative to Global Attention, Luong et al. \cite{b40} propose Local Attention. By setting a context window, only a small part of the source language can be generated when generating the context vector, and some irrelevant information can be filtered to reduce the amount of calculation. Bahdanau et al. \cite{b41} successfully apply the attention mechanism to the field of machine translation, greatly improving the translation effect. Wu et al. \cite{b35} present a Siamese attention architecture, and embed the attention mechanism into spatial gated recurrent units to selectively propagate relevant features and memorize their spatial dependencies through the network. Wu et al. \cite{b42} propose a deep attention-based spatially recursive model that can learn to attend to critical object parts and encode them into spatially expressive representations. which is composed of two-stream CNN layers, bilinear pooling, and spatial recursive encoding with attention, is end-to-end trainable to serve as the part detector and feature extractor whereby relevant features are localized, extracted, and encoded spatially for recognition purpose. Wu et al. \cite{b43} propose a network which is further augmented with 3D part alignment module to learn local features through Soft Attention module. These attended features are statistically aggregated to yield identity-discriminative representations.

Self attention is a special case of the attention mechanism. Different from the above, the self-attention mechanism only needs a sequence to calculate the representation. Through the interaction of the internal elements of the sequence, the structural information inside the sequence is learned to obtain better representation learning. It has achieved very good results in many NLP tasks, including machine translation \cite{b44}, reading comprehension \cite{b45}, statement representation, text summary \cite{b46}, language comprehension \cite{b47} and other tasks. In this paper, we add the self-attention mechanism to our pipeline framework.

\section{Our Two Frameworks}\label{Approach}

\subsection{Task Definition}

Single-relation factoid questions (i.e., simple questions) can be answered by a single fact of the KBs.
The formal definition is as follows. KB \( \{s_i, r_i,o_i\} \)  is a set of triples, where \( s_i \) and  \( o_i \)  are the subject entity and object entity,  \( r_i \)  is the relation,  \( (s_i, r_i,o_i) \)  corresponds to one fact. For the purpose of answering question \emph{q} formulated in natural language, we need to find a triple  \( (s, r,o) \) , where \( s \) and \( o \) correspond to the subject and predicate in the question \emph{q},  then \( o \) is the answer to the question \emph{q}.

Therefore as long as we find the corresponding subject and predicate, we can turn question into a structured query to obtain the answer.


\subsection{Our Pipeline Framework}\label{pipeline framework}
\subsubsection{Model Description}
\begin{figure}[h]
\centering
\includegraphics[width=1.0\textwidth, height=0.4\textheight]{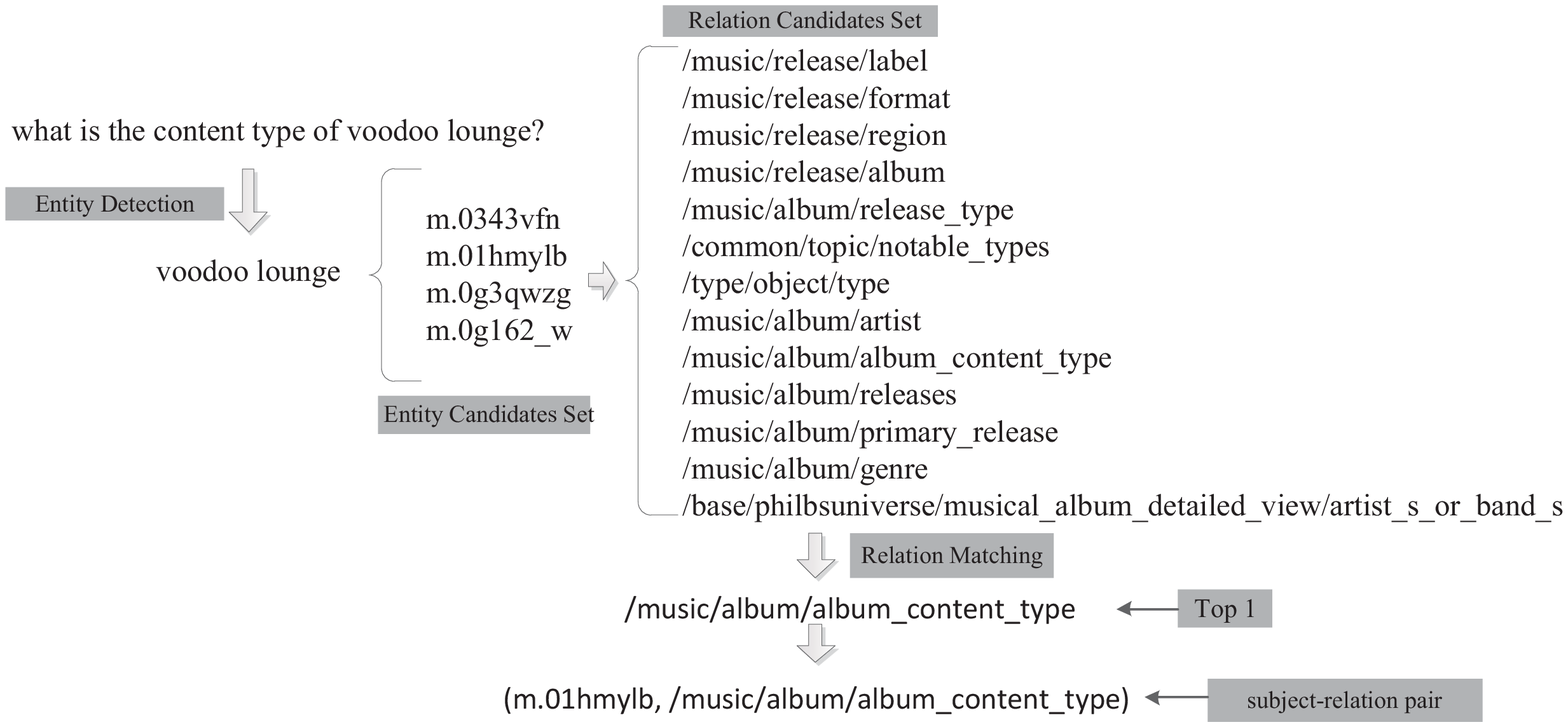}
\caption{Example of simple QA process: First, the entity is retrieved through entity detection according to the given question, and the entity corresponding to the fact in the KB is matched to obtain an entity candidate set. Second, according to the entities in the entity candidate set, the related relations are found, and the relation candidate set is formed. At last, according to the matching of the relation and the question, the relation with the highest matching degree is obtained, and the best entity-relation pair is obtained accordingly.\label{simple_qa}}
\end{figure}

In this section, we divide the QA task into two subtasks: entity detection and relation detection.
\begin{enumerate}
  \item We generate entity candidate set \emph{E} by entity detection.
  \item Based on entity candidate set \emph{E}, we obtain all of the relations associated with the entity candidates.
  \item Then we calculate the semantic similarity between the relation and the question by semantic matching model, and take the relation \emph{r} with the highest matching score as the result to relation detection.
  \item Finally we select the corresponding triple as the answer based on \emph{r}.
\end{enumerate}
As shown in the Figure \ref{simple_qa}, after getting entity candidate set,``/music/album/album\_conten\\t\_type" is the highest matching relation with the question. At this time, we get the entity-relation pair. According the entity-relation pair, we can find the triple (m.01hmylb, /music/album/album\_content\_type, m.06vw6v) as the fact to answer the given question.

\subsubsection{Entity Detection}
Following the traditional approach, we first conduct entity recognition to mark out the words that belong to the entity in the question. Like the mainstream method, we treat it as a sequence labeling task. We train a bidirectional LSTM \cite{b48,b49} network to detect entity text in the question.


\begin{figure}[h]
\centering
\includegraphics[width=0.8\textwidth]{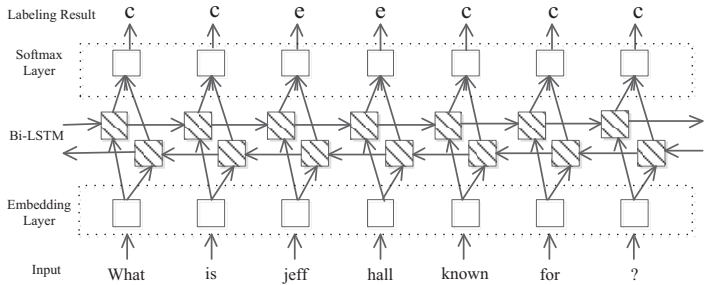}
\caption{The architecture of entity recognition: We use Bi-LSTM for entity recognition to extract the entity text.\label{entity_detection}}
\end{figure}

As shown in Figure \ref{entity_detection}, the character ``e'' is the entity text, and ``c'' is the context text. the words that belong to the entity text are marked e, and the words that do not belong to the entity text are marked c. The fragment of entity text refers to the sequence of words corresponding to each consecutive segment e.

Because some results of entity recognition may not be completely correct, we have to employ some strategies to remedy it, therefore we borrow the method proposed by Ture et al. \cite{b50}. Based on the result of entity recognition, we obtain the fragment of entity text. Then we construct the entity candidate set through the following process.

\begin{enumerate}
  \item We find out all entities in FB2M whose alias exactly equal to the fragment of entity text, and then form entity candidate set  {\emph{E}}. If there is no exact match entity, go to the next step.
  \item The 1-gram, 2-gram and 3-gram from each fragment of entity text are extracted. If a tuple is a subset of another tuple, we should keep the long one and discard the short one. Then we can form a set of n-gram \( G \) .
  \item We search the entities based on n-gram and form an entity candidate set  \emph{E}. Using  Equation \ref{weight of entity} to calculate the weight of the entity. Then we take the entity with the weight equal to the highest score as the result of entity detection.
\end{enumerate}

\begin{equation}\label{weight of entity}
 \begin{split}
score_i= \frac {N_i}{ L_iC_i}
\end{split}
\end{equation}

Where \( N_i \) is the number of words in \( G \), \( L_i \)  is the number of words contained in the entity tag of the retrieved entity, and \( C_i \) is the number of entities in the \( E \).

\subsubsection{Relation Detection}
\begin{figure}[h]
\centering
\includegraphics[width=0.8\textwidth]{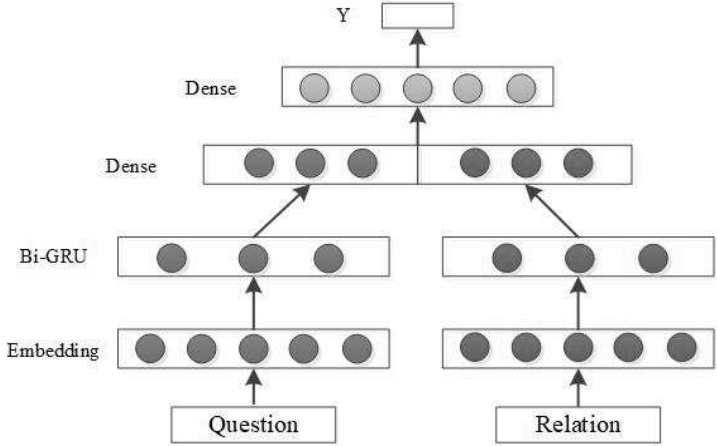}
\caption{The network structure of semantic matching model.\label{semantic matching model}}
\end{figure}

In this subtask, our core mission is to measure the semantic similarity between the relations and questions. Therefore, we design a semantic matching model. We take the matching task as a binary classification problem, matching or not matching. First we design a binary classification network. Instead of using the classification result directly, we use the value of the output layer as the matching score. As shown in Figure \ref{semantic matching model}, in this process, we input the sequence text for the question and the relation, and get the word vector representation of each word through the Embedding layer. And then we get the semantic representation of the question and relation through Bi-GRU \cite{b51,b52}, and stitch them into a vector. As well as through a fully connected layer the final score can be obtained in the output layer. By the way, we use sigmoid as the activation function of the output layer.

\subsubsection{Context Information Of Pipeline Framework}\label{context in pipe}
In this paper, we explore to improve the resolution of the entities with the same name by entities' context information. Here context information includes the entities' out-degree and notable type.

According to whether or not the context information is combined, the selection of the entity-relation pair is divided into three different algorithms: no context information (Pipeline QA without context information i.e., P-QA) as shown in Algorithm 1, combined with out-degree information (Pipeline QA with out-degree information i.e., P-QA-Out) as shown in Algorithm 2 and combined with type information (Pipeline QA with type information i.e., P-QA-Type) as shown in Algorithm 3.

\begin{enumerate}
  \item In Algorithm 1, after obtaining the entity candidate set \emph{$E(e_1,e_2,...,e_m)$} and the relation \emph{r} with the highest matching degree, the entity with the relation \emph{r} can be selected as the result of the final entity detection.

\begin{algorithm}
  \renewcommand{\algorithmicrequire}{\textbf{Input:}}
  \renewcommand{\algorithmicensure}{\textbf{Output:}}
  \caption{Prediction algorithm without context information(P-QA)}
  \begin{algorithmic}[1]
    \REQUIRE
        question \emph{Q}\\
        entity recognition model \emph{EM} \\
        relation matching model \emph{RM} \\
    \ENSURE "entity-relation" dual group \emph{(e,r)} \\
    \STATE We use the model \emph{EM} to perform entity identification on the question \emph{Q}, and then the entity text \emph{P} is obtained. \\
    \STATE The entity candidate set \emph{E} is generated using entity retrieval algorithm. \\
    \STATE According to the entity candidate set \emph{E}, we form a relation candidate set \emph{R}. \\
    \STATE We calculate the matching score of the question \emph{Q} and all relations according to the relation matching model. \\
    \STATE According to the relation matching result, the relation \emph{r} with the highest score is selected as the result of the relation matching. \\
    \STATE We find the corresponding entity \emph{e} from the entity candidate set \emph{E} according to \emph{r}. \\
    \STATE The prediction results i.e.,entity-relation \emph{(e,r)} dual group is output. \\
  \end{algorithmic}
\end{algorithm}
\label{pipeline without context}

  \item However, sometimes there are multiple entities in E that have the same relation r, therefore here the re-sorting of the entities is performed using the out-degree information to generate the final result.

In Algorithm 2, The out-degree of the entity \emph{e} is the number of triples in the KB in which \emph{e} is the subject. We try to rank the entity candidate set based on the out-degree. The greater the number of out-degree of an entity, the greater the scope of its association in the KB.

\begin{algorithm}
  \renewcommand{\algorithmicrequire}{\textbf{Input:}}
  \renewcommand{\algorithmicensure}{\textbf{Output:}}
  \caption{Prediction algorithm with out-degree information(P-QA-Out)}
  \begin{algorithmic}[1]
    \REQUIRE
        entity candidate set \emph{$E(e_1,e_2,...,e_m)$}\\
        relation \emph{r} \\
    \ENSURE "entity-relation" dual group \emph{(e,r)} \\
    \STATE The entity candidate set \emph{E} is pruned according to \emph{r}, and an entity candidate subset \emph{$E^{'}(e^{'}_{1},e^{'}_{2},...,e^{'}_{k})$} with the relation \emph{r} can be generated. \\
   \STATE We calculate the out-degree of all entities in the candidate set \emph{$E^{'}$}, and \emph{$O_i(o^{'}_{1},o^{'}_{2},...,o^{'}_{k})$} can be generated. \\
   \STATE Then we sort \emph{$E^{'}(e^{'}_{1},e^{'}_{2},...,e^{'}_{k})$} in descending order according to \emph{$ O_i$}. \\
   \STATE The entity with the highest degree of \emph{e} is selected as the result of entity detection. \\
   \STATE The prediction results i.e.,entity-relation \emph{(e,r)} dual group is output. \\
  \end{algorithmic}
\end{algorithm}

\item In Algorithm 3, we also try to use the type information of the entity to distinguish the entities with same name. The notable type of the entity in FreeBase is a simple atomic label that indicates what the entity is notable for \cite{b53}.  Freebase was acquired by Google in 2010 and officially shut down in 2016. Its data was migrated to Wikidata. Since Freebase's online API is closed, it is impossible to get the notable type information directly. The Freebase data dumps can be downloaded in an N-Triples RDF format.

We extract the notable type information from the dump files. There are 1275 kind of notable types in 2 million entities of FB2M. Then we use the same network structure like relation detection to calculate the matching score between questions and notable types. And we try to improve the accuracy of entity recognition by ranking candidate entities based on the matching scores.

Here, the matching score of the entity type information and the question is recorded as \emph{$S_t$}, and the matching score of the relation and the question is recorded as \emph{$S_r$}. Then the two are added to obtain the "entity-relation" binary group and the matching score of question \emph{S} as shown in Equation \ref{matching score of question}. The final screening of the results is achieved by scoring and sorting the entity-relation pairs.

\begin{equation}\label{matching score of question}
S=S_t+S_r
\end{equation}

\begin{algorithm}
  \renewcommand{\algorithmicrequire}{\textbf{Input:}}
  \renewcommand{\algorithmicensure}{\textbf{Output:}}
  \caption{Prediction algorithm with entity type(P-QA-Type)}
  \begin{algorithmic}[1]
    \REQUIRE
        question \emph{Q} \\
        entity candidate set \emph{$E(e_1,e_2,...,e_m)$}\\
        relation candidate set \emph{$R(r_1,r_2,...,r_n)$}\\
    \ENSURE "entity-relation" dual group \emph{(e,r)} \\
    \STATE The entity type is extracted and a type list \emph{$T(t_1,t_2,...,t_m)$} is generated corresponding to the entity candidate set \emph{E}. \\
    \STATE We calculate the semantic matching score between the entity type \emph{T} and the question \emph{Q}, \emph{$S_{t}(s^{1}_{t},s^{2}_{t},...,s^{m}_{t})$} is generated. \\
    \STATE We calculate the semantic matching score between the relation \emph{R} and the question \emph{Q}, \emph{$S_{r}(s^{1}_{r},s^{2}_{r},...,s^{m}_{r})$} and entity-relation binary list \emph{$P(p_1,p_2,...,p_m)$} are generated. \\
    \STATE According to \emph{$S_t$} and \emph{$S_r$}, we add them one by one according to the list \emph{P} to generate \emph{$S(s_1,s_2,...,s_m)$}. \\
   \STATE We sort the list \emph{P} in descending order according to the comprehensive score \emph{S}. \\
   \STATE The prediction results i.e.,entity-relation \emph{(e,r)} dual group is output. \\
    \end{algorithmic}
\end{algorithm}

\end{enumerate}

\subsubsection{Loss Function Of Our Pipeline Framework}
In our pipeline framework, we use two different loss functions.
  \begin{enumerate}
  \item In entity recognition, we use categorical\_crossentropy (i.e., Equation \ref{categorical_crossentropy}) as loss function.
\begin{equation}\label{categorical_crossentropy}
 C=-\frac {1}{n}\sum_{x}[yln{a}+(1-y)ln(1-a)]
\end{equation}

Where $y$ is the expected output and $a$ is the actual output. This function has two properties: (1) non-negative. (2) when the actual output a is close to the expected output y, the loss function is close to 0. (For instance, $y$=0, $a$\~{}0; $y$=1, $a$\~{}1, the loss function is both close to 0.). In addition, this function also can overcome the problem that the variance cost function update weight is too slow.

  \item In relation matching and in the matching of questions and type information which use Algorithm 3, The Equation \ref{categorical_crossentropy_2} is as the loss function. which is a little different from Equation \ref{categorical_crossentropy}, and all characters represent the same meaning.

\begin{equation}\label{categorical_crossentropy_2}
 C=-\frac {1}{n}[yln{a}+(1-y)ln(1-a)]
\end{equation}
  \end{enumerate}

\subsection{Our End-to-end Framework}\label{end-whole-framework}
\subsubsection{Model Description}

In this section, we explore an end-to-end framework for answering simple questions. Without entity recognition, the entity \emph{e} is retrieved directly by retrieving the n-gram tuple generated by question \emph{q}. through entering the question and each fact into this model, we can get the \emph{score(fact)} of each fact. And then we sort the facts by these scores, therefore we can get the final answer according to the sort result. This approach aims to build a more versatile QA model. In addition, we also study the use of self-attention mechanisms and joint representation learning, and build an end-to-end joint learning model that combines self-attention mechanisms.

\begin{figure}[h]
\centering
\includegraphics[width=1.0\textwidth, height=0.5\textheight]{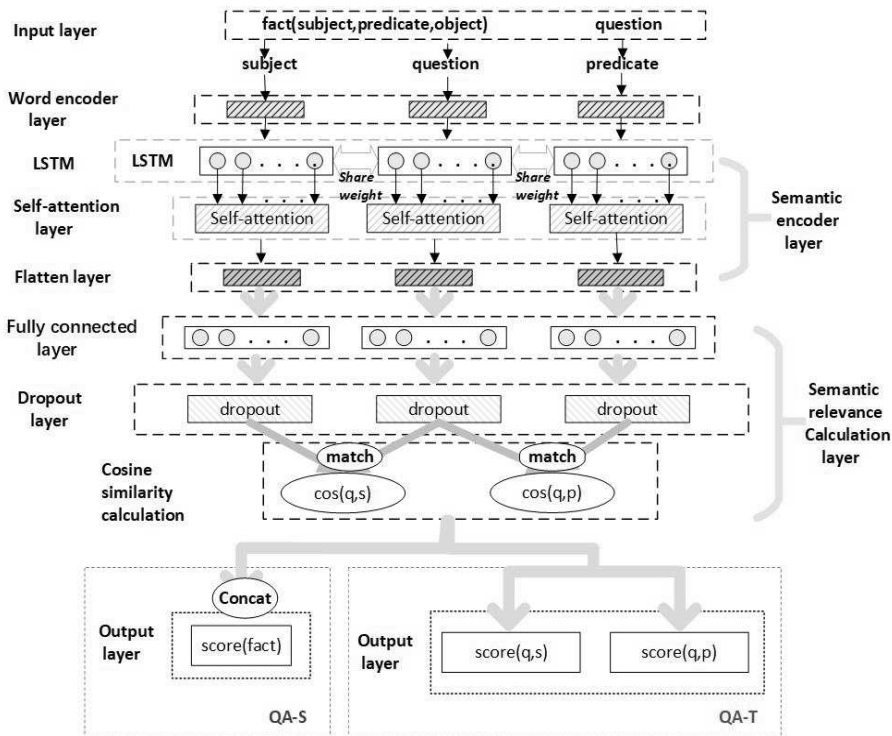}
\caption{Visualization of the whole model. Subject, question, predicate are sequentially respectively encoded by this model, producing their word vector representation, and then they are encoded into their semantic vector representation. Question-subject pair and question-predicate pair are scored using a cosine similarity between their representing vectors, then we can get score(fact) in two different ways.\label{end-whole-model}}
\end{figure}

As shown in Figure \ref{end-whole-model}, our entire model consists of 5 major layers (or 9 small layers). The first 4 parts get the cosine similarity score of question-entity pair and question-relation pair respectively. Then we score the fact according to two different methods. We call them the automatic QA method based on weight sharing (we will call it QA-S below) and the automatic QA method based on weight sharing and multi-task (we will call it QA-T below). The shared part of the two methods in the model is further elaborated in Section \ref{end-to-end framework}, followed by a description of QA-S in Section \ref{description of QA-S} and a description of QA-T in Section \ref{description of QA-T}.

\subsubsection{Description Of Each Layer}\label{end-to-end framework}

\begin{enumerate}
  \item In Input layer: We enter the fact \emph{f(subject, predicate, object)} and question \emph{q} to the model, retrieve the subject, the predicate and send them with the question to the model for the next module to process.

  \item In Word encoder layer: We encode the retrieved subject, question, and predicate to get their word vector representation, using multi-granular coding method. and the network structure is shown in Figure \ref{word-encoding}. This encoding combines word-level encoding and char-level encoding to obtain the semantic vector of the word. For the word \emph{w \((c_1,c_2,...,c_n) \)}, where $c_i$ is the ith character of the word \emph{w}, the corresponding word vector is \emph{$\vec{v_w}$}. First, the character sequence \((c_1,c_2,...,c_n) \) is input to the GRU network, with the last hidden layer state \emph{$\vec{h_i}$} as the word \emph{w}'s semantic representation \emph{$\vec{v_c}$} of the char-level,  then \emph{$\vec{v_w}$} and \emph{$\vec{v_c}$} are concatenated as the semantic vector \emph{v}\([\vec v_w,\vec v_c] \) of the word \emph{w}.

\begin{figure}[h]
\centering
\includegraphics[width=0.6\textwidth,height=0.4\textheight]{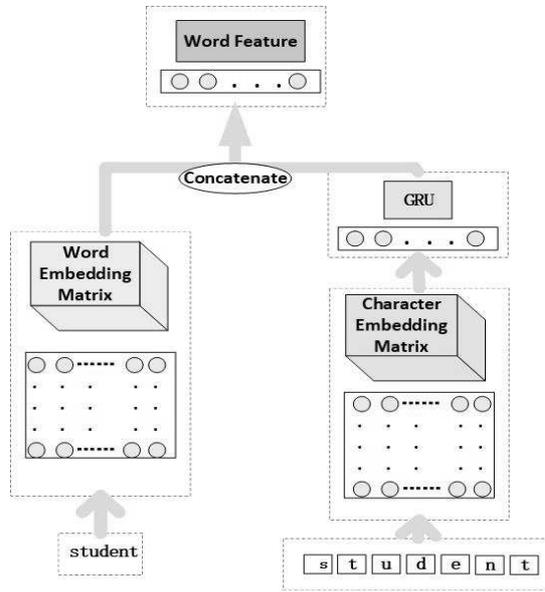}
\caption{The network structure of word encoder layer.\label{word-encoding}}
\end{figure}

\item Semantic encoder layer: It consists of three parts: LSTM, Self-attention layer and Flatten layer.
  \begin{enumerate}
  \item In LSTM: We send the semantic vector sequences of the subject, the question, and the predicate obtained by the word encoder layer into the same LSTM recurrent neural network for semantic encoding. In this part, subjects, questions and predicates share weights, which makes the processing more concise.

  \item In Self-attention layer: We add the self-attention mechanism to the semantic vector representation of the subject, the question and the predicate, which can capture the long-distance interdependent features in the sentence. Since the final hidden layer state \emph{$\vec h_i$} of LSTM is used as the semantic encoding of subjects, questions, and predicates, and the previous t-1 states have been discarded. However, the hidden state set \emph{h \(\lceil \vec h_1,\vec h_2,...,\vec h_n \rceil\)} also corresponds to the semantic information of the current sentence. Therefore, by adding a self-attention mechanism, the model can fully calculate the internal structure of the sentence by calculating the attention between the hidden layer state sets h. And the model can generate a sentence-level embedding matrix, which makes the semantic representation vector more accurately reflect the contents of the subject, the question, and the predicate.

  \item In Flatten layer: We compress the semantic representation matrix vectors of the subject, the question, and the predicate obtained in the previous layer into one-dimensional semantic representation vectors for convenient processing.
  \end{enumerate}

  \item Semantic relevance calculation layer: It consists of Fully connected layer, Dropout layer and Cosine similarity calculation.
    \begin{enumerate}
  \item In Fully connected layer: All the semantic vectors obtained are respectively made into a nonlinear transformation to obtain the probability distribution of the subject, the predicate and the question. The activation function used here is Rectified Linear Unit (i.e., ReLU) as Equation \ref{end-to-end ReLU}.

\begin{equation}\label{end-to-end ReLU}
d=f(W_{a}x+b)
\end{equation}

Where $W_a$ is the weight coefficient of the full connection, $b$ is the bias term, and $f$ is the ReLU activation function as shown in Equation \ref{end-to-end f}.

\begin{equation}\label{end-to-end f}
f(x)=max(0,x)
\end{equation}

  \item In Dropout layer: It mainly prevents the over-fitting by temporarily discarding the neural network unit from the network according to a certain probability. The Dropout concept was proposed by Hinton et al. \cite{b54} to solve the over-fitting problem in deep image classification. Dropout refers to setting the output of each hidden neural unit to 0 with probability p as shown in Figure \ref{Dropout layer diagram.}, therefore the neural unit whose output is set to 0 will not participate in the subsequent forward transmission of the network, and will not participate in the reverse. In addition Dropout can reduce the complex co-adaptation between neurons and force the network to learn more general characteristics \cite{b55}.

\begin{figure}[h]
      \centering
      \subfloat{
        \includegraphics[width = .45\textwidth]{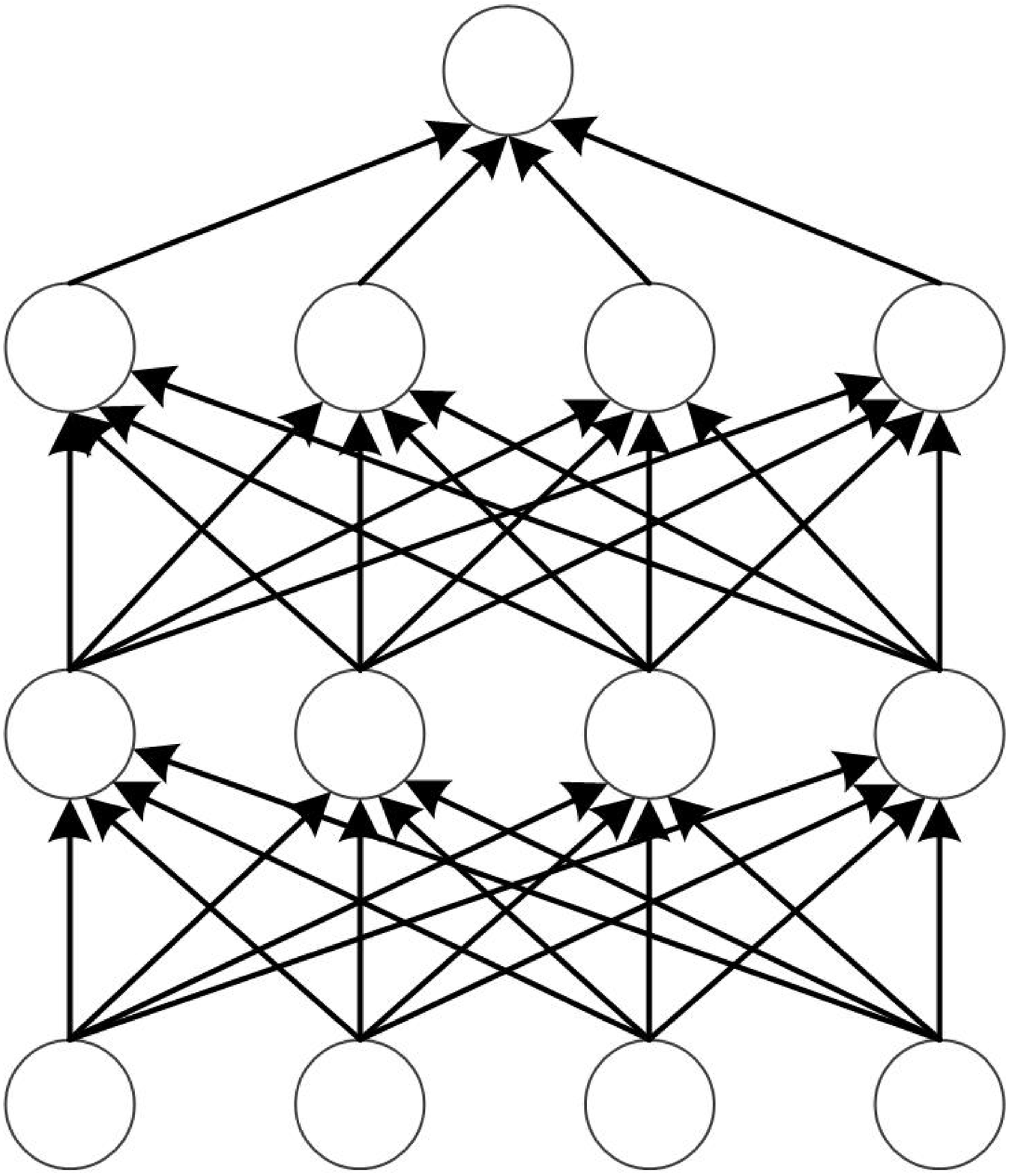}
        \label{sub1}
      }
      \qquad
      \subfloat{
        \includegraphics[width = .45\textwidth]{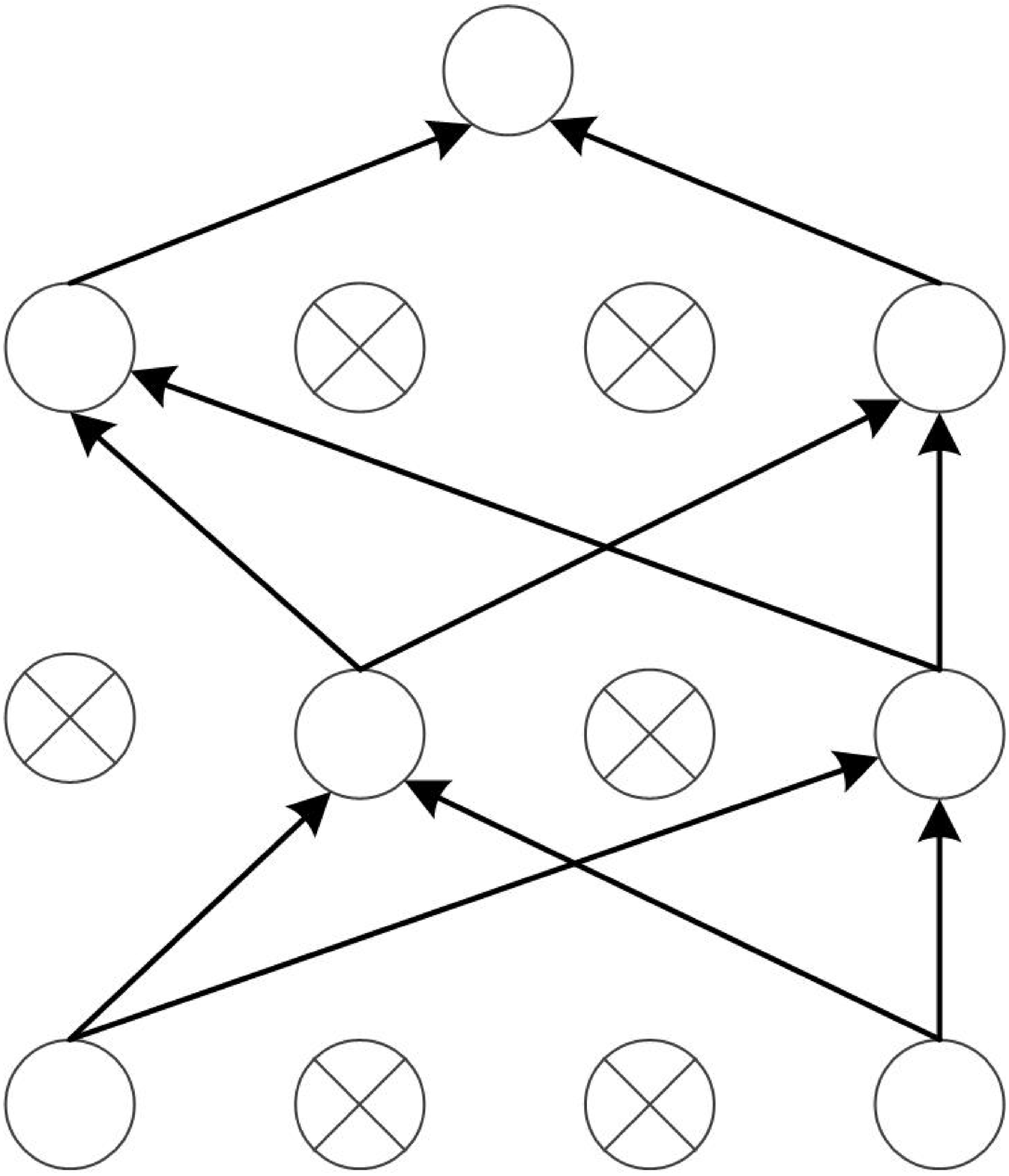}
        \label{sub2}
      }
      \caption{Dropout layer diagram.}\label{Dropout layer diagram.}
    \end{figure}

  \item In Cosine similarity calculation: We calculate the matching score $score(q,s)$ between the question and the subject and the matching score $score(q,p)$ of the question and the predicate through Equation \ref{end-to-end score_q s} and Equation \ref{end-to-end score_q p}, respectively. The cosine similarity calculation formula is shown in Equation \ref{end-to-end cos}.

\begin{equation}\label{end-to-end score_q s}
score(q,s)=cos(\vec q,\vec s)
\end{equation}

\begin{equation}\label{end-to-end score_q p}
score(q,p)=cos(\vec q,\vec p)
\end{equation}

\begin{equation}\label{end-to-end cos}
cos(a,b)=\frac {\vec a * \vec b}{\| \vec a\| \|\vec b\|}
\end{equation}

  \end{enumerate}

\end{enumerate}

The above are the shared parts of the two scoring methods. Below we will describe the two different scoring methods in details.

\subsubsection{Description Of QA-S}\label{description of QA-S}

Through the first eight parts of the model, we can get the matching scores of the question-subject and the question-predicate pair, Based on the concatenation of them, we can get the score(fact) through Equation \ref{QA-S socre}

\begin{equation}\label{QA-S socre}
score= W[S_{q,s};S_{q,p}]
\end{equation}

where $W$ is the weight matrix, $ S_{q,s}$ is the matching score of question-subject pair, and $ S_{q,p}$ is the matching score of question-predicate pair.

\subsubsection{Description Of QA-T}\label{description of QA-T}
Multi-task learning has been successful in machine learning, such as NLP, voice recognition, and image annotation. It improves the generalization of the model by using shared representations and learning multiple tasks at the same time, and learning domain knowledge in related tasks during the training process. Rei \cite{b56} propose a sequence tagging framework with auxiliary tasks to improve the performance of the main task, which motivate the model to learn the general patterns of semantics and syntax by training the context of the predicted words. Segaard et al. \cite{b57} propose a multi-level sharing model, which outputs low-level tasks in low-level networks and outputs advanced tasks in high-level networks, that is, design sharing patterns according to the semantic level of tasks. Hashimoto et al. \cite{b58} construct an end-to-end NLP framework for multi-task joint learning, which can be used to handle tasks such as part-of-speech tagging, chunking, dependency parsing, and textual entailment.

Different from QA-S, we do not concatenate the cosine similarity of the question-subject pair and the question-predicate pair, but separately, and linearly transform them according Equation \ref{score question-subject} and Equation \ref{score question-predicate} to obtain the score of subject s and predicate p respectively.

\begin{equation}\label{score question-subject}
score\langle q,s \rangle= w_a \cos{(\vec q,\vec s)}
\end{equation}

\begin{equation}\label{score question-predicate}
score\langle q,p \rangle=w_b \cos{(\vec q,\vec p)}
\end{equation}

where $w_a$, $w_b$ are the linear conversion coefficients of the subject and predicate, respectively.

The above is the full introduction of our end-to-end framework. Two different methods are used to score the facts and then we will evaluate our models and methods in Section \ref{Experiments}.

\subsubsection{Context Information Of End-to-end Framework}\label{context information of end-to-end}

Similar to Section \ref{context in pipe}, we explore the role of context information in the end-to-end framework for QA accuracy in this section.

\begin{enumerate}
  \item We use two different methods to incorporate entity type information, as follows:
  \begin{enumerate}
  \item In the end-to-end framework using method QA-T, the strategy adopted is to integrate the entity type information into the entity tag encoding process corresponding to the entity, that is, to convert the entity tag into an entity tag with entity type information. For example, for the question "\textbf{\emph{how was germany released ?}}", the subject is "\textbf{\emph{germany}}", its corresponding target entity is \textbf{\emph{m.017hzy7}}, the entity type is "\textbf{\emph{musical recording}}", then we incorporate the entity type information into the entity tag. therefore we can get the new entity label "\textbf{\emph{musical recording germany}}" corresponding to \textbf{\emph{m.017hzy7}}. In addition, for the two methods that do not use or use the self-attention mechanism, they are recorded as \textbf{\emph{QA-T-wt}} and \textbf{\emph{QA-T-swt}}, respectively.

  \item In the same framework with method QA-T as above, a strategy different from the above is adopted. In the Section \ref{context in pipe}, a matching model of the question and the entity type information is specifically established in order to calculate the matching score of the question and the relation. Therefore drawing on this idea, the matching of the question with the entity type information is an independent task. That is, on the output of QA-T, the matching score of the entity type information and the question is added. As an auxiliary task, it can not only improve the model to mine the entity type information in the question, but also promote the matching of the entity, the relation and the question. According to the use and non-use of the self-attention mechanism, recorded two methods as \textbf{\emph{QA-T-mwt}} and \textbf{\emph{QA-T-mwst}} respectively.
  \end{enumerate}

  \item In addition to using entity type information to help distinguish entities of the same name, we can also use the out-degree information to sort the answers with the highest matching, in order to further sort and filter the answers. This section verifies the effect of the out-degree from the "Word", "Word + Self Attention" based on the end-to-end framework with QA-S, and from the end-to-end framework with QA-T that combines the entity type information. The specific experimental results will be shown in Section \ref{experiment of end-to-end}.

\end{enumerate}

\subsubsection{Loss Function Of Our End-to-end Framework}
In our end-to-end framework, We use a total of three loss functions, in which all characters have the same meaning, where \emph{$s^+$} is the positive sample of the entity corresponding to the question \emph{q}, i.e., the target entity. \emph{$t^+$} is the entity type information of \emph{$s^+$}, \emph{$s^-$} is the negative sample of the entity corresponding to \emph{q}, \emph{$t^-$} is the entity type information of \emph{$s^-$}; \emph{$p^+$} is the positive sample of the relation corresponding to \emph{q}, i.e., the target relation, and \emph{$p^-$} is the negative sample of the relation corresponding to \emph{q}, \emph{$\gamma$} is a hyperparameter. $S_s$, $S_p$ and $S_t$ are the matching scores corresponding to the question-subject pair, the question-predicate pair, and the question-type pair respectively.

Through the established model in Section \ref{end-to-end framework}, the semantic matching scores of the question-subject pair, the question-predicate pair can be separately calculated. In the training process, the positive entity sample $s^+$ and the positive relation sample $r^+$ of the question q are input, and a group is also input. Corresponding negative entity sample $s^-$ and negative relation  sample $r^-$ . Under normal circumstances, the matching score of the question and the positive sample should be as much as possible than the matching score of the question and the negative sample.

(1)For the end-to-end framework with method QA-S, the loss function is shown in Equation \ref{QA-S-lossfunction}. Because the matching scores of questions and subjects and predicates have been merged into a score at the output layer of the model, that is, the matching scores of the question and the "subject-predicate" correspond to $S(q,s^+,p^+)$ and $S(q,s^-,p^-)$.

\begin{equation}\label{QA-S-lossfunction}
loss=max(0,S(q,s^-,p^-)+\gamma-S(q,s^+,p^+))
\end{equation}

(2)For the end-to-end framework with method QA-T, the loss function is shown in Equation \ref{QA-T-lossfunction}. The value of the formula consists of two parts: the entity matching loss value and the relation matching loss value.

\begin{equation}\label{QA-T-lossfunction}
 loss=\sum_{(q,s^+,p^+)}(max(0,S_s(q,s^-)-S_s(q,s^+)+\gamma
 )+max(0, S_p(q,p^-)-S_p(q,p^+)+ \gamma))
\end{equation}

The purpose of using the loss function is not to focus on the specific numerical value of the matching scores, but to focus on the difference of matching scores between the positive and negative samples and the question q, therefore strengthen the model's ability to distinguish between positive and negative samples, which can make the model learn better. The hyperparameter $\gamma$ is defined, therefore the similarity value between the question and the positive sample can exceed the range of the similarity value between the question and the negative sample, as shown in Equation \ref{QA-T-gamma}.

\begin{equation}\label{QA-T-gamma}
S_s(q,s^+)-S_s(q,s^-)\geq \gamma
\end{equation}

(3)In addition, since in the Section \ref{context information of end-to-end} the matching score of the entity type and the question is added in the second method of incorporating the type information, the loss function is different from the above, and the matching score of the question and the entity type is added based on the Equation \ref{QA-T-lossfunction}, as shown in Equation \ref{loss function of multi-end}.

\begin{equation}\label{loss function of multi-end}
\begin{aligned}
loss = \sum_{q,s^+,p^+,t^+}(max(0,S_s(q,s^-)+\gamma-S_s(q,s^+))
+max(0,s_p(q,p^-)+\gamma-S_p(q,p^+)) \\
+max(0,s_-t(q,t^-)+\gamma-S_t(q,t^+)))
 \end{aligned}
\end{equation}


\section{Experiments}\label{Experiments}
\subsection{Dataset}

we utilize SimpleQuestions dataset, which is released by Bordes et al. \cite{b26} and consists of 108442 questions written in natural language. It is constructed according corresponding facts in Freebase. The facts format as (subject, predicate, object). According to the original data division ratio, there are 75910 training data, 21687 test data, and 10845 validation data. This dataset also provides two subsets of Freebase: FB2M and FB5M. They are represented as sets of triples. We take FB2M as background KB, it includes 2 million entities and 6701 relations. In addition, we also use Freebase data dumps (22 GB compressed, 250 GB uncompressed) to extract entities' notable type information.

\subsection{Experiment Configuration and Results Of Our Pipeline Framework}

\subsubsection{Generation Of The Training Set}\label{generation of training data set}
\begin{figure}[h]
\centering
\includegraphics[width=1.0\textwidth ,height=0.5\textheight]{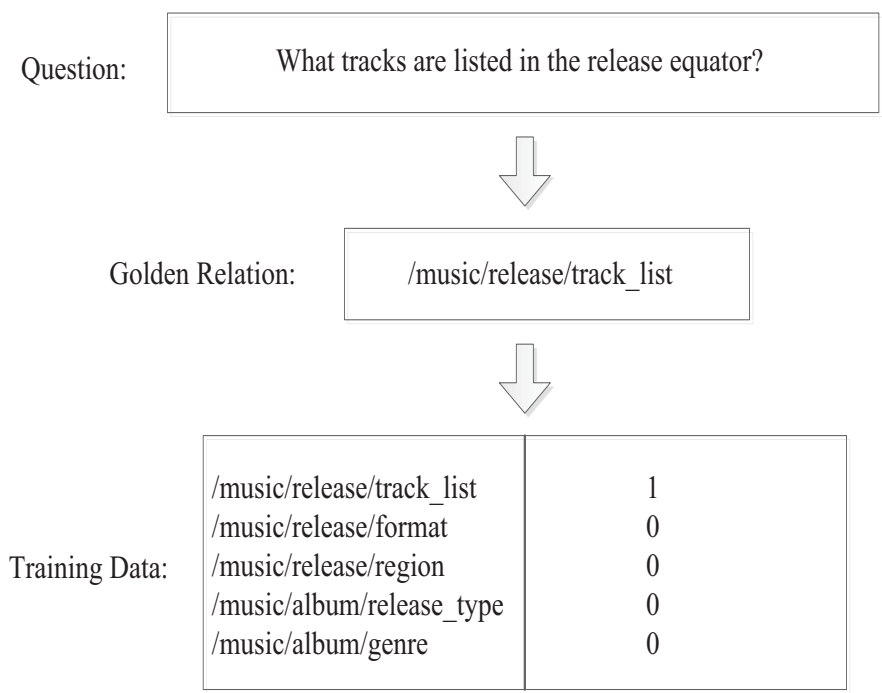}
\caption{Example of generating relation training data.\label{generation of relation training data}}
\end{figure}

\begin{enumerate}
  \item \textbf{Construction of the positive training set:}
  We construct a positive sample training set using the training data from the SimpleQuestions dataset mentioned above.
  \item \textbf{Construction of the negative training set:}
  \begin{enumerate}
  \item \textbf{Generation of the Entity training set \emph{$E^-$\(({e_1,e_2,...,e_{n-1}}\))}:} In order to generate annotation data for entity identification, it is necessary to solve the string matching of the question \emph{q} with the corresponding entity \emph{e} to separate the entity text from the context text. By analyzing the experimental data, the entity text in some questions \emph{q} is not strictly matched with the target entity \emph{e}, and there are problems such as singular and plural numbers, misspellings, etc. Therefore it is impossible to perform only an exact match between the question \emph{q} and the entity \emph{e}. in order to properly flag all questions, we design the following algorithm:
 \begin{enumerate}
  \item First, the question's word sequence \emph{$q(w_1,w_2,...,w_n)$} is input to generate a 1-(n-1) tuple of \emph{q}, which is denoted as \emph{$g(g_1,g_2,...,g_n)$}.
  \item Then the following operations are performed on each entity in the KB: We calculate the edit distance \emph{$l(l_1,l_2,...,l_n)$} of entity \emph{e} and each element in the \emph{g}, and the tuple g corresponding to \emph{l} is taken as the matching segment.
  \item Finally, according to the question \emph{q} marked by \emph{g}, all the word sequences corresponding to \emph{q} are marked as '\emph{e}', and other words are marked as '\emph{c}', and all word sequences marked as '\emph{e}' constitute the Entity training set \emph{$E^-$\(({e_1,e_2,...,e_{n-1}}\))}.
  \end{enumerate}

  \item \textbf{Generation of the Relation training set \emph{(q,r,tag)}:} According to the first word of each relation, we divide all relations into 89 major categories, representing 89 domains. For example, in the domain of music, several relation examples are shown in Table \ref{relation_example in music domain}. We generate training data in units of questions.

 \begin{enumerate}
  \item For each question \emph{q} and the triple  \( (e,r,o) \), we first determine the domain \emph{D} according to the golden relation \emph{r}.
  \item Then we get all the relations \emph{R} of the domain \emph{D}. We generate pairs formatted in the form of  \( (q, R_i,tag) \) where tag is equal to 0 or 1.
  \item As illustrated in Figure \ref{generation of relation training data}, the relation corresponding to the question belongs to the music domain. Thus, we pair all the relations belonging to the music domain with questions and form corresponding tags. Where if relation is the target relation, then tag=1, otherwise tag=0. The Relation training set is constructed.
  \end{enumerate}

  \end{enumerate}
\end{enumerate}

In addition, we copy positive cases three times in order to reduce or avoid the impact of data imbalance. Because in the construction of training data, the proportion of negative samples generated is much larger than positive samples.

\begin{table}[!h]
\begin{center}
  \centering
  \caption{Several relation examples in music domain.}
  \label{relation_example in music domain}
  \begin{footnotesize}
   \begin{threeparttable}
\begin{tabular}{|c|}
\hline
\textbf{/music/live\_album/concert\_tour} \\
\textbf{/music/composition/compose} \\
\textbf{/music/release/label} \\
\textbf{/music/album\_content\_type/albums} \\
\textbf{/music/recording/producer} \\
\textbf{/music/genre/parent\_genre} \\
\textbf{/music/artist/concert\_tours} \\
\textbf{/music/album/compositions} \\
\textbf{......} \\
\hline
\end{tabular}
\end{threeparttable}
\end{footnotesize}
\end{center}
\end{table}

\subsubsection{Training Settings}

The model word embeddings are initialized with the 300-dimensional pre-trained vectors provided by Glove \cite{b59}. We update network weights by using the Adam \cite{b60} optimizer with learning rate 0.001. The hidden layers of Bi-LSTM and Bi-GRU have size 100. In the semantic matching model, Dropout is set to 0.1.

\subsubsection{Initial Preparation Work}

\begin{enumerate}
  \item All entitiy names are divided into words, and 1-gram, 2-gram and 3-gram are generated. Then we build an inverted index \emph{I} that maps all n-grams to the entity's alias text.
  \item The entity's notable type information is extracted from Freebase data dumps.
  \item According to the training set, the question text should be labeled and an entity recognition training set is generated.
\end{enumerate}

\subsubsection{Experimental Results and Discussions}

Only if the entity and the relation are correctly predicted, the question \emph{q} is considered being answered correctly. Therefore we use accuracy of entity-relation pair to measure the final QA results.

\begin{enumerate}
  \item \textbf{Entity Detection: }The experiment result shows that the accuracy rate of entity recognition is 82. 2\%, of which 31.7\% of entities cannot be uniquely identified because one name or alias may corresponds to multiple entities. In this case, the corresponding entity cannot be uniquely identified by the name or alias alone. Besides, 17.8\% of entities are not fully labeled correctly. This part needs to retrieve entities based on the result of entity recognition and form entity candidate sets.
  \item \textbf{Relation Detection: }We generate a test data set for the model of relation detection based on the test set. For each question, we take the relations associated with its golden entity as the relation candidates. Then we use the model to find the highest matching relation. After testing, the accuracy rate of relation matching can reach 90.1\%.
  \item \textbf{Combination with out-degree and notable type information: }As shown in Table \ref{pipeline with type and degree_1}, after adding the out-degree information (method P-QA-Out), the accuracy rate increases from 65.5\% to 66.6\%, an increase of 0.9\%, and the "Error with same label entity" has decreased by 1.1\%. When the entity type information (method P-QA-Type) is added, the accuracy rate is increased by 0.7\%, and the proportion of the error type "Error with same label entity" is also reduced by 0.7\%.

\begin{table}[htbp]
\caption{Comparison of results after integrating type information and out-degree in our pipeline framework. “Error with same label entity” means that the entity label in the prediction result is the same as the label of the target entity, but the entity ID is different.}
\renewcommand\arraystretch{1.5}
\begin{center}
\begin{tabular}{|l|c|c|}
\hline
Approach & Error with same label entity &  Accuracy\\
\hline
P-QA & 15.3\% &	65.5\%  \\
\hline
 P-QA-Out & 14.2\% &	\textbf{\emph{66.6\%}}  \\
\hline
P-QA-Type & 14.6\% &	66.2\%  \\
\hline
P-QA-Out-Type  & 14.3\% &	66.5\%  \\
\hline
P-QA-Type-Out  & 14.5\% &	66.3\%  \\
\hline
\end{tabular}
\label{pipeline with type and degree_1}
\end{center}
\end{table}

At the same time, we also explore the combination of out-degree information and entity type information. Based on the method P-QA-Out, the answers are reordered by combining the entity type information to obtain the result of the method P-QA-Out-Type. It can be seen that the final accuracy is reduced by 0.1\%. In addition, in the method P-QA-Type, the answers are reordered by combining the out-degree information to obtain the result of the method P-QA-Type-Out. Compare method P-QA-Type-Out with method P-QA-Type, the final accuracy rate increased by 0.1\%.

In the related work of combining entity type information in the QA, Dai et al. \cite{b28} transformed the matching problem of entity type and relation into a multi-classification task, that is, the training model classifies the question and determines the corresponding entity type in the question. On the Pipeline technology route, no strict comparisons have been found on the SimpleQuestions dataset to distinguish entities with the same name based on out-degree information.

On the Pipeline technology route, in the entity identification process, Dai et al. \cite{b28} use the Bi-GRU-CRF framework for entity identification and use Bi-GRU networks for relation matching. Yin et al.\cite{b29} use the Bi-LSTM-CRF framework for entity recognition, and then combine the character matching algorithm to classify the entity candidate set. In the fact selection process, entity label matching and relation matching are performed using a convolutional neural network (CNN) based on char-level and word-level, respectively. In this paper, we use the results of entity linking provided by Yin et al. \cite{b29}, combined with the entity detection model, and the experiments are carried out according to the method in Section \ref{pipeline framework}, and the results are shown in Table \ref{pipeline with type and degree_2}.

\begin{table}[htbp]
\caption{Comparison of results after integrating entity type information and out-degree in our pipeline framework using the entity linking result from Yin et al. \cite{b29}.}
\renewcommand\arraystretch{1.5}
\begin{center}
\begin{tabular}{|l|c|c|}
\hline
Approach & Error with same label entity &  Accuracy\\
\hline
PC-QA & 9.1\% &	70.4\%  \\
\hline
PC-QA-Out & 8.3\% &	71.2\%  \\
\hline
PC-QA-Type & 8.2\% &	71.3\%  \\
\hline
PC-QA-Out-Type  & 8.0\% &	71.5\%  \\
\hline
PC-QA-Type-Out  & \textbf{7.9}\% &	\textbf{71.6}\%  \\
\hline
\end{tabular}
\label{pipeline with type and degree_2}
\end{center}
\end{table}

As can be seen from Table \ref{pipeline with type and degree_2}, compared with Table \ref{pipeline with type and degree_1}, the overall result of the experiment is greatly improved, and the effect is more obvious when using the entity type information and out-degree information. After combining them, the experimental results are found to increase by 1.1\% and 1.2\%, respectively. Therefore, it can be seen that entity identification and entity detection have a great influence on the results of subsequent processes, and there is a problem of error transmission.

The experimental results show that the pipeline-based technical route can effectively complete the QA task. The type information and out-degree information can improve the QA results. In this process, multiple modules need to be built, and corresponding training data sets must be constructed separately. The whole process is complex and costly, and the error transmission problem needs to be solved.

\end{enumerate}

\subsection{Experiment Configuration and Results Of Our End-to-end Framework}

\subsubsection{Generation Of The Training Set}

\begin{enumerate}
  \item \textbf{Construction of the positive training set:}
      We use the same method as in Section \ref{generation of training data set} to construct the positive sample training set.
  \item \textbf{Construction of the negative training set:}
      As there are thousands of subjects and predicates in the KBs, it is impossible to use all subjects except the true subject and all predicates except the true predicate as the training data. For generating the training data set efficiently, some of the most valuable samples need to be selected to train the model.
      Therefore we will generate the training data set in the following three steps:
      \begin{enumerate}
      \item \textbf{Generation of the dictionary \emph{$ D_{rr} $}:} We calculate the edit distance \emph{L} of each predicate in the Predicate set P and other predicates except it (which are expressed as \emph{p\'} ), then all \emph{p\'} are sorted in ascending order according to \emph{L}, thereby the dictionary \emph{ $ D_{rr} $ } is generated.

      \item \textbf{Generation of the Subject training set \emph{$S^-$\(({s_1,s_2,...,s_{n-1}}\))}:} For question \emph{q}, the target subject is denoted as \emph{s} with a label, the predicate is \emph{p}, and the generated subject candidate set is \emph{S\( ({s_1,s_2,...,s_n}\))}. First, we remove the subject which has the same label with the target subject from the subject candidate set \emph{S}. Then the remaining subjects constitute the Subject training set \emph{$S^-$\( ({s_1,s_2,...,s_i}\) )} whose size is i. If i is less than 5, randomly select ( 5 - i ) subjects from the subject candidate set \emph{S}. Each time a training subject is randomly selected from \emph{$S^-$\( (s_1,s_2,...,s_5)\)}.
      \item \textbf{Generation of the Predicate training set \emph{$ P^-$\((p_1,p_2,...,p_{n-1})\)}:} The true predicate p is removed from the candidate predicate set \emph{P} (which is composed of all predicates related to the subject \emph{s}), then the remaining predicates constitute the Predicate training set \emph{$P^-$} whose size is j. If j is less than 50, it is sequentially added from the dictionary \emph{$D_{rr}$} until equal to 50. Random sampling of \emph{$P^-$} without returning each time a negative sample is selected.
      \end{enumerate}
\end{enumerate}

\subsubsection{Training Settings}
We evaluate our method on the SimpleQuestions dataset which contains N = 21.687 questions and the corresponding triples. For each question we follow the procedure described in Section \ref{end-whole-framework} to find whether adding char-level encoding or adding self-attention mechanism will help the matching of question and target fact, and which scoring method can get the answer of the question more accurately.

\subsubsection{Experimental Results and Discussions}\label{experiment of end-to-end}

\begin{enumerate}
  \item \textbf{Combination with char-level encoding and self-attention mechanisms: }
\begin{table}[htbp]
\caption{Comparison of results between QA-S and QA-T with char-level encoding and self-attention mechanisms. ``Word'' means that only word-level encoding is used in the word encoder layer, and char-level encoding is not incorporated. ``Word + Self Attention'' refers to the application of a self-attention mechanism at the semantic encoder level based on the ``Word'' coding. ``Word+Character'' is a multi-granular encoding method that uses word-level encoding and char-level encoding in the word encoder layer. ``Word+Character+Self Attention'' refers to the application of the self-attention mechanism in the semantic encoder layer based on ``Word+Character''.}
\renewcommand\arraystretch{1.5}
\begin{center}
	\begin{tabular}{|l|c|c|}
        \hline
		\multirow{2}*{Approach} & \multicolumn{2}{|c|}{Accuracy} \\
		\cline{2-3}
		~ & model QA-S & model QA-T\\
		\hline
Word & 56.1\% & 67.6\% \\
		\hline
Word+Self Attention & 67.3\% & \textbf{\emph{70.7\%}} \\
		\hline
Word+Character & 59.4\% & 68.8\% \\
		\hline
Word+Character+Self Attention & 64.0\% & 69.8\% \\
		\hline
\end{tabular}
\label{end-to-end with cha and self att}
\end{center}
\end{table}

First, we can see from the Table \ref{end-to-end with cha and self att} that the multi-tasking end-to-end model based on weight sharing (QA-T) is better than the end-to-end model based on weight sharing (QA-S). The following sections will compare the two aspects of char-level encoding and self-attention mechanisms.
 \begin{enumerate}
      \item It can be seen that after the character information is integrated into the word encoder layer (i.e., from ``Word'' to ``Word+Character''), it has increased by 3.3\% in the model QA-S, from 56.1\% to 59.4\%. In the model QA-T, it increased from 67.6\% to 68.8\%, an increase of 1.2\%.

      \item After adding the self-attention mechanism to the semantic encoder layer (i.e., from ``Word'' to ``Word+Self Attention''), we can see that the result of the model QA-S has improved from 56.1\% to 63.7\%, which improved by 7.6\%; and the accuracy of the model QA-T has increased from 67.6\% to 70.7\%, an increase of 3.1\%.
      \item After adding the self-attention mechanism to the word encoder layer based on the embedding of character information (i.e., from ``Word+Character'' to ``Word+Character+Self Attention''), the result of the model QA-S is from 59.4\% to 64.0\%, an increase of 4.6\%; the result of the model QA-T has increased by 1.0\%, from 68.8\% to 69.8\%.
      \item After the multi-granularity encoding method is adopted on the basis of the semantic encoder layer adding the self-attention mechanism (i.e., from ``Word + Self Attention'' to ``Word+Character+Self Attention''). the result of the model QA-S has increased by 0.3\%. However, the result of Model  QA=T has decreased by 0.9\%.
      \end{enumerate}

From the overall comparison of model QA-S and model QA-T, we can find that the latter have a bigger improvement compared to the former. Under the same conditions, the addition of the self-attention mechanism can get improvement in both models, and the effect is more obvious in the model QA-S. The effect of the word encoder layer incorporating words' char-level encoding is reflected in the model QA-S, Which can alleviate OOV problems and enhance the semantic representation of each word. In the model QA-T, The embedding of char-level encoding can increase the accuracy of the model by 1.2\% without adding a self-attention mechanism. However, after adding the attention mechanism, the character information has a negative impact on the model's effect, down by 0.9\%.

The reason why the accuracy of the model QA-T has decreased may be that the character vector is initialized by a random vector and is continuously tuned during the training of the model. However, the data in the training data set is limited, therefore it results in a low quality of the character vector. When each element interacts in the self-attention mechanism, the noise is amplified, resulting in a decrease in the final result.

The experimental results show that the self-attention mechanism can better mine the internal structure of questions, entities and relations, and achieve better results without additional use of character information. The following is an example to explore the impact of the self-attention mechanism on the semantic encoder layer. For example, the question \textbf{\emph{``Which genre of album is harder ..... faster?''}}, the corresponding subject is \textbf{\emph{``harder ..... faster''}}, and the corresponding relation is \textbf{\emph{``music/album/genre''}}. The attention weight matrix corresponding to the question and relation is shown in Figure \ref{question_weight} and Figure \ref{relation_weight} respectively. The darker the color, the larger the weight.

\begin{figure}
\begin{minipage}[t]{0.5\linewidth}
\centering
\includegraphics[width=1.0\textwidth,height=0.4\textheight]{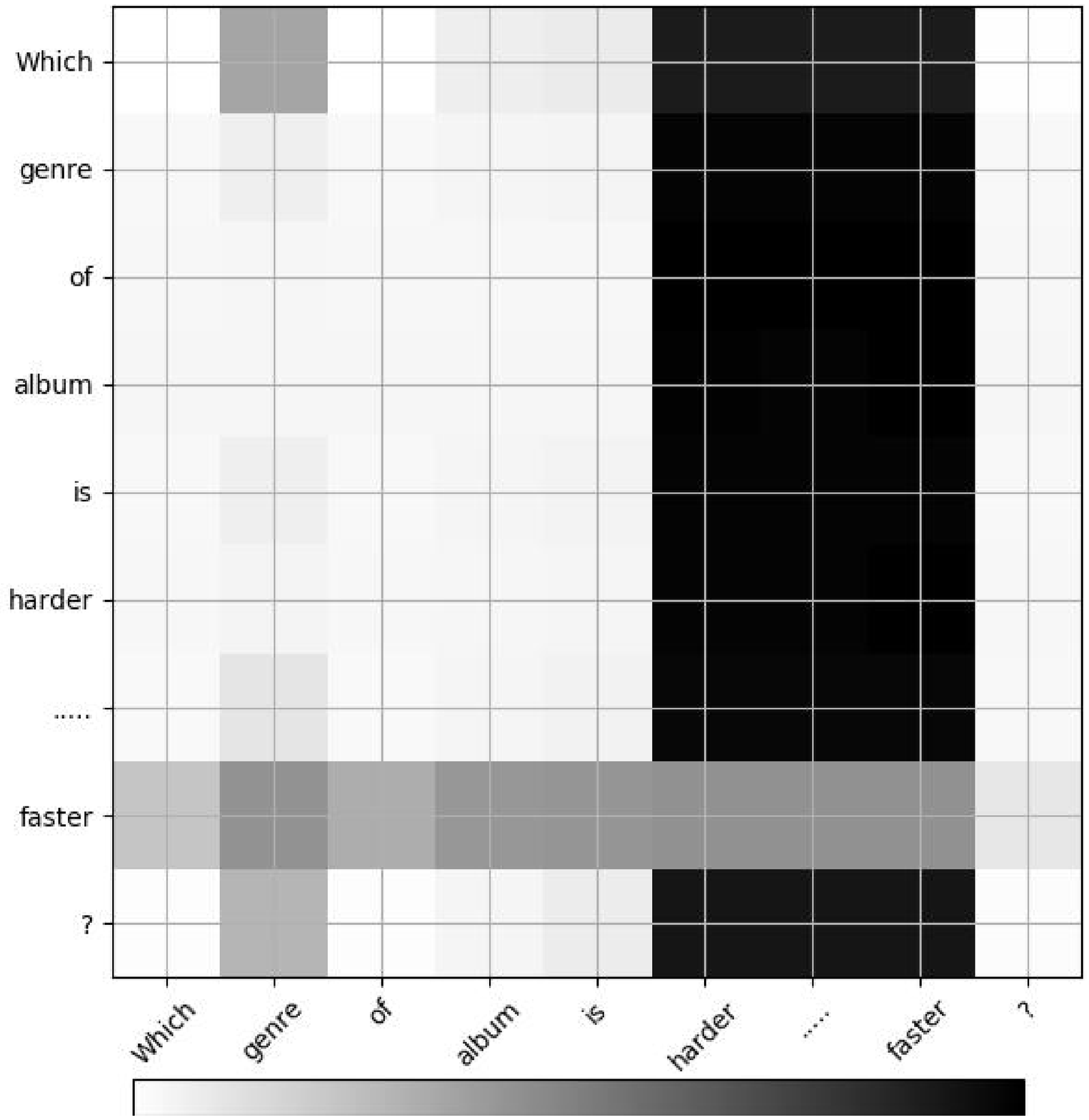}
\caption{Question weight distribution \\with example ``Which genre of albu\\-m is harder ..... faster?''.}
\label{question_weight}
\end{minipage}%
\begin{minipage}[t]{0.5\linewidth}
\centering
\includegraphics[width=0.8\textwidth,height=0.25\textheight]{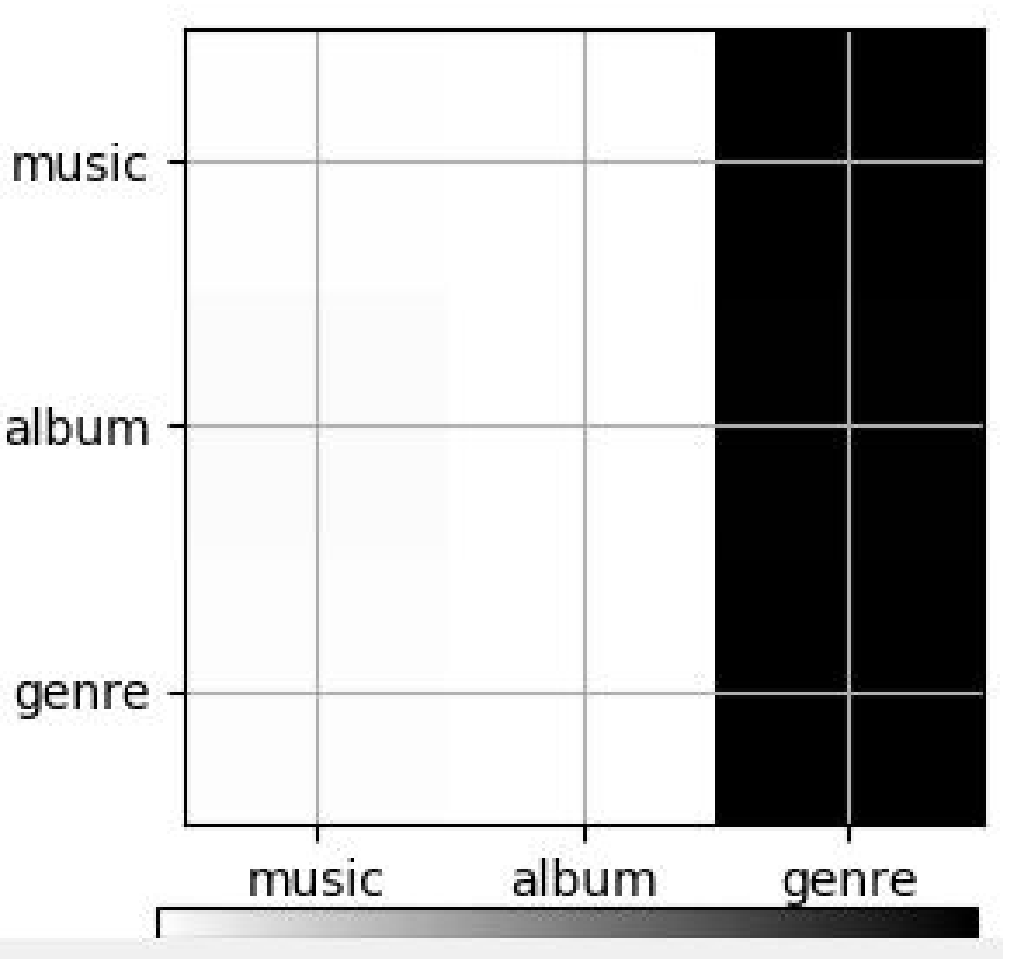}
\caption{Relation weight distribution \\with example. As can been seen in this figure, `genre' has the largest weight in the relation `/music/album/genre'.}
\label{relation_weight}
\end{minipage}
\end{figure}

It can be seen from Figure \ref{question_weight} that in the attention matrix of the question, the three words ``harder ..... faster'' are given a large weight, corresponding to the subject in the question. In addition, the word ``genre'' is given a relatively large weight in the question. It shows that the internal structure of the sentence is well studied by using the self-attention mechanism, and the subject information and predicate information are more concerned in the question.

In the attention matrix corresponding to the relation ``music/album/genre'' as show\\n in Figure \ref{relation_weight}, the weight of ``genre'' is large, which in turn corresponds to the predicate information ``genre'' in the above question. This shows that the use of the self-attention mechanism not only captures the important and discriminative information in the relation, but also echoes the predicate information in the question.

\item \textbf{Combination with out-degree and notable type information: }

\begin{table}[htbp]
\caption{Comparison of results after integrating entity type information.}
\renewcommand\arraystretch{1.5}
\begin{center}
\begin{tabular}{|l|c|c|c|c|c|}
\hline
Approach & Indistinguishable & Ambiguity & Wrong\_S & Wrong\_P & Accuracy \\
\hline
QA-T-w & 3.4\% &	5.0\% & 5.5\% & 7.4\% & 67.3\% \\
\hline
QA-T-wt & 3.5\% &	4.0\% & 5.8\% & 7.1\% & \textbf{\emph{68.3\%}} \\
\hline
QA-T-mwt & 3.5\% &	3.9\% & 5.4\% & 7.3\% & \textbf{\emph{68.6\%}} \\
\hline
QA-T-ws  &	3.7\% &	4.8\% & 3.1\% & 6.8\% & 70.3\% \\
\hline
QA-T-wst  &	3.5\% &	3.8\% & 6.9\% & 6.2\% & 68.3\% \\
\hline
QA-T-mwst  & 3.5\% & 3.8\% & 3.0\% & 6.8\% & \textbf{\emph{71.5\%}} \\
\hline
\end{tabular}
\label{end-to-end with type}
\end{center}
\end{table}

In Table \ref{end-to-end with type}, QA-T-w represents the method of applying char-level encoding in the multi-task end-to-end model based on weight sharing(QA-T), and QA-T-ws represents the application of self-attention mechanism on the basis of QA-Tw. “Wrong\_S” stands for “Wrong Subject Mention” and “Wrong\_P” stands for “Wrong Predicate”.

As can be seen from Table \ref{end-to-end with type}, after the entity tag combines the entity type information, the accuracy of the model QA-T-wt is 1.0\% higher than that of the model QA-Tw, and the case of "Ambiguity" is reduced from 5.0\% to 4.0\%. Therefore, when the self-attention mechanism is not added, the introduction of entity type information enriches the coding information of the entity and enhances the resolving power of the model for the case of "Ambiguity".

After adding the self-attention mechanism, the QA-T-wst model has no improvement in the final accuracy of the model compared with the QA-T-ws model, and the error rate of the entity ("Wrong\_S") has increased, but its “Ambiguity” situation has dropped from 4.8\% to 3.8\%.

From the methods QA-T-mwt and QA-T-mwst, it can be seen that the matching of the entity type and the question is a separate task, the effect of the model has been significantly improved. Method QA-T-mwt is 1.3\% higher than QA-T-w. Compared with the model QA-T-ws, the accuracy rate of the model QA-T-mwst has increased from 70.3\% to 71.5\%, increased by 1.2\%, and the "Ambiiguity" error rate decreased by 1.0\%, while the entity label error rate ("Wrong\_S" ") fell by 0.1\%.

In addition to using entity type information to help distinguish entities with the same name, we also use the out-degree information. Here the effect of it is shown in Table \ref{end-to-end with out-degree}. It can be seen that when the out-degree information is added, the results of each method will have an improvement of about 0.4\%, the highest result of the experiment in this paper is 71.8\% on the SimpleQuestions dataset using the QA-T-mwst method. Therefore, combining the entity information and the out-degree information can effectively improve the effect of QA.

\begin{table}
[htbp]
\caption{Comparison of the results of the model before and after adding out-degree information.}
\renewcommand\arraystretch{1.5}
\begin{center}
	\begin{tabular}{|c|c|c|}
		\hline
		\multirow{2}*{Approach} & \multicolumn{2}{|c|}{Accuracy} \\
		\cline{2-3}
		~ & unsorted & sorted\\
		\hline
QA-T-w & 67.3\% & 67.6\% \\
		\hline
QA-T-ws & 70.3\% & 70.7\% \\
		\hline
QA-T-mwst & 71.5\% & \textbf{\emph{71.8\%}} \\
		\hline
\end{tabular}
\label{end-to-end with out-degree}
\end{center}
\end{table}

\begin{table}[htbp]
\caption{Comparison of end-to-end framework research results. Note: [*] indicates that the result is a model result when FB5M is the background KB.}
\renewcommand\arraystretch{1.5}
\begin{center}
\begin{tabular}{|p{4.5cm}|p{2.5cm}|}
\hline
Approach	  & Accuracy  \\
\hline
Bordes et $ al^{ [26] } $  &	62.7\%   \\
\hline
Yin et $ al^{ [29] } $  & 68.3\%  \\
\hline
Dai et $ al^{ [28] } $  & $62.6\%^{[*]}$  \\
\hline
Golub and He $ al^{ [33] } $  & 70.9\%  \\
\hline
Lukovnikov et $ al^{ [34] } $ & \textbf{\emph{71.2\%}}  \\
\hline
Our approach  & \textbf{\emph{71.8\%}}  \\
\hline
\end{tabular}
\label{accuracy of many end-to-end framework}
\end{center}
\end{table}

\begin{table}[htbp]
\caption{Comparison of model runtime and accuracy}
\renewcommand\arraystretch{1.5}
\begin{center}
	\begin{tabular}{|l|c|c|c|c|c|}
\hline
Approach	  & GPU  & Batch Size & Epoch & Time & Approach   \\
\hline
Lukovnikov et $ al^{ [34] } $   & Titan X & 100 &50 & 48h & 71.2\%  \\
\hline
Our Approach & GTX 1080Ti & 512 & 40 & \textbf{\emph{4h}} & \textbf{\emph{71.8\%}} \\
\hline
\end{tabular}
\label{Model running time and accuracy comparison}
\end{center}
\end{table}

\end{enumerate}

The relevant research results of the currently known end-to-end framework on the SimpleQuestions dataset are shown in Table \ref{accuracy of many end-to-end framework}. In the end-to-end route, the work of Lukovnikov et al. \cite{b34} is currently known to be the best, and our paper uses a much less than their time to achieve a slightly better result than theirs, an increase of 0.6\%. The best result of this paper is obtained by incorporating entity type information on a multi-task end-to-end framework using word-level encoding combined with a self-attention mechanism. Lukovnikov et al. \cite{b34} also utilized character information in the encoding of questions. They use CNN to semantically model the character sequence of each word, and then the character semantic information is incorporated into the semantic representation of the corresponding word, which is used to alleviate the OOV question and also at the cost of partial time cost.

\section{Conclusions and Future Work}\label{Conclusion}

In this paper, we compare the impact of the entity's out-degree, notable type information on answering single-relation factoid questions based on KBs through two different frameworks: a pipeline framework and an end-to-end framework. In the former framework, we divide this task into two subtasks: entity detection and relation detection. In the latter framework, we combine char-level encoding and self-attention mechanisms, using method of sharing weights and method of multitasking on fact selection. As the experimental results show, combining context information can improve the accuracy of QA in both frameworks. It helps to distinguish ambiguity of the entity with the same name. In addition, we find there are some ambiguities that cannot be resolved in limited context information. In practical applications, the combination of the questioner's identity information (user profile information) and more context information may solve the problem to some extent. Second, the accuracy of QA also can benefit from the adding of char-level encoding information and self-attention mechanisms. It helps achieve a more accurate semantic representation of the entity, to find the better match of fact, therefore questions can be answered with the better answer.

In future work,, we will improve the accuracy of QA from the following aspects:

\begin{enumerate}
  \item In this paper, CNN is used to semantically encode questions, entities and relations, map them to the same semantic space, and then get the matching scores between semantic vectors by measuring the similarity of semantic vectors. However, there is no interaction between the character information of questions, entities and relations. Therefore in the subsequent research, we will try to combine more deep semantic matching models.
  \item We mainly explore the use of two types of context information in this paper: entity type and out-degree information. In the future research work, we can consider introducing more context information. For example, in Freebase, the entity also has the attribute ``common / topic / description'', which is a holistic description of the entity. In addition, we can try to use the context information of the target entity as feedback to correct the results of QA.
  \item It can be seen from the experimental process that the proportion of the negative sample has a great influence on the training result of the model within a certain range. In the follow-up study, the negative sampling method can be further explored to further improve the training efficiency and the effect of the QA model.
\end{enumerate}

\end{document}